# Hierarchical POMDP Controller Optimization by Likelihood Maximization

**Marc Toussaint**
Computer Science
TU Berlin
Berlin, Germany
mtoussai@cs.tu-berlin.de

**Laurent Charlin**
Computer Science
University of Toronto
Toronto, Ontario, Canada
lcharlin@cs.toronto.edu

**Pascal Poupart**
Computer Science
University of Waterloo
Waterloo, Ontario, Canada
ppoupart@cs.uwaterloo.ca

## Abstract

Planning can often be simplified by decomposing the task into smaller tasks arranged hierarchically. Charlin et al. [4] recently showed that the hierarchy discovery problem can be framed as a non-convex optimization problem. However, the inherent computational difficulty of solving such an optimization problem makes it hard to scale to real-world problems. In another line of research, Toussaint et al. [18] developed a method to solve planning problems by maximum-likelihood estimation. In this paper, we show how the hierarchy discovery problem in partially observable domains can be tackled using a similar maximum likelihood approach. Our technique first transforms the problem into a dynamic Bayesian network through which a hierarchical structure can naturally be discovered while optimizing the policy. Experimental results demonstrate that this approach scales better than previous techniques based on non-convex optimization.

## 1 Introduction

Planning in partially observable domains is notoriously difficult. However, many planning tasks naturally decompose into subtasks that may be arranged hierarchically. For instance, the design of a soccer playing robot is often decomposed into low-level skills such as intercepting the ball, controlling the ball, passing the ball, etc. [16]. Similarly, prompting systems that assist older adults with activities of daily living (e.g., handwashing [8]) can be naturally decomposed into subtasks for each step of an activity. When a decomposition or hierarchy is known a priori, several approaches have demonstrated that planning can be simplified and performed faster [13, 7]. However, the hierarchy is not always known or easy to specify, and the optimal policy may only decompose *approximately*. To that effect, Charlin et al. [4] showed how a hierarchy can be discovered automatically by formulating the planning problem as a non-convex quartically constrained optimization problem with variables corresponding to the parameters of the policy, including its hierarchical structure. Unfortunately, the inherent computational difficulty of solving this optimization problem prevents the approach from scaling to real-world problems. Furthermore, it is not clear that automated hierarchy discovery simplifies planning since the space of policies remains the same.

We propose an alternative approach that demonstrates that hierarchy discovery (i) can be done efficiently and (ii) performs a policy search with a different bias than non-hierarchical approaches that is advantageous when there exists good hierarchical policies. The approach combines Murphy and Paskin's [10] factored encoding of hierarchical structures (see also [17]) into a dynamic Bayesian network (DBN) with Toussaint et al.'s [18] maximum-likelihood estimation technique for policy optimization. More precisely, we encode POMDPs with hierarchical controllers into a DBN in such a way that the policy and hierarchy parameters are entries of some conditional probability tables. We also consider factored policies that are more general than hierarchical controllers. The policy and hierarchy parameters are optimized with the expectation-maximization (EM) algorithm [5]. Since each iteration of EM essentially consists of inference queries, the approach scales easily.

Sect. 2 briefly introduces partially observable Markov decision processes, controllers and policy optimization by maximum likelihood estimation. Sect. 3 reviews previous work on hierarchical modeling and how to use a dynamic Bayesian network to encode a hierarchical structure. Sect. 4 describes our proposed approach, which combines a dynamic Bayesian network encoding with maximum likelihood estimation to simultane-ously optimize a hierarchy and the controller. Sect. 5 demonstrates the scalability of the proposed approach on benchmark problems. Finally, Sect. 6 summarizes the paper and discusses future work.

## 2 Background

Throughout the paper we denote random variables by upper case letters (e.g., $X$), values of random variables by their corresponding lower case letters (e.g., $x \in dom(X)$) and sets of values by upper case letters with math calligraphy (e.g., $\mathcal{X} = \{x_1, x_2, x_3\}$). We now review POMDPs (Sect. 2.1), how to represent policies as finite state controllers (Sect. 2.2) and how to optimize bounded controllers (Sect. 2.3).

### 2.1 POMDPs

Partially observable Markov decision processes (POMDPs) provide a natural and principled framework for planning. A POMDP can be formally defined by a tuple $\langle \mathcal{S}, \mathcal{A}, \mathcal{O}, p_s, p_{s'|as}, p_{o'|s'a}, r_{as} \rangle$ where $\mathcal{S}$ is the set of states $s$, $\mathcal{A}$ is the set of actions $a$, $\mathcal{O}$ is the set of observations $o$, $p_s = \Pr(S_0 = s)$ is the initial state distribution (a.k.a. initial belief), $p_{s'|as} = \Pr(S_{t+1} = s' \mid A_t = a, S_t = s)$ is the transition distribution, $p_{o'|s'a} = \Pr(O_{t+1} = o' \mid S_{t+1} = s', A_t = a)$ is the observation distribution and $r_{as} = R(A_t = a, S_t = s)$ is the reward function. Throughout the paper, it is assumed that $\mathcal{S}$, $\mathcal{A}$ and $\mathcal{O}$ are finite and discrete. The goal is to select actions to maximize the rewards. At any point in time, the information available to select the next action consists of the history of past actions and observations. Hence a policy $\pi$ is defined as a mapping from histories to actions. However, since histories grow with time, it is common practice to summarize histories with a fixed-length sufficient statistic such as the belief distribution $b_s = \Pr(S = s)$, which corresponds to the state distribution (conditioned on the history of past actions and observations). The belief distribution $b$ can be updated at each time step, based on the action $a$ taken and the observation $o'$ made according to Bayes' theorem: $b_{s'}^{ao'} = k \sum_s b_s p_{s'|as} p_{o'|s'a}$ (k is a normalization constant). Policies can then be defined as mappings from beliefs to actions (e.g., $\pi(b) = a$). The value $V^\pi(b)$ of a policy $\pi$ starting in belief $b$ is measured by the discounted sum of expected rewards: $V^\pi(b) = \sum_t \gamma^t E_{b_t|\pi}[r_{\pi(b_t)b_t}]$ where $r_{ab} = \sum_s b_s r_{as}$. An optimal policy $\pi^*$ is a policy with the highest value $V^*$ for all beliefs: $V^*(b) \geq V^\pi \forall \pi, b$. The optimal value function also satisfies Bellman's equation: $V^*(b) = \max_a r_{ab} + \sum_{o'} p_{o'|ab} V^*(b^{ao'})$ where $p_{o'|ab} = \sum_{ss'} b_s p_{s'|as} p_{o'|s'a}$.

### 2.2 Finite State Controllers

A convenient representation for an important class of policies consists of finite state controllers [6]. Instead of using beliefs as sufficient statistics of histories, the idea is to use a finite internal memory to retain relevant bits of information from histories. Each configuration of this memory can be thought of as a node in a finite state controller, where nodes select actions to be executed and edges indicate how to update nodes based on the observations received. A controller with a finite set $\mathcal{N}$ of nodes $n$ can encode a stochastic policy $\pi$ with three distributions: $\Pr(N_0 = n) = p_n$ (initial node distribution), $\Pr(A_t = a \mid N_t = n) = p_{a|n}$ (action selection distribution) and $\Pr(N_{t+1} = n' \mid N_t = n, O_{t+1} = o') = p_{n'|no'}$ (successor node distribution). Such a policy can be executed by starting in a node $n$ sampled from $p_n$, executing an action $a$ sampled from $p_{a|n}$, receiving observation $o'$, transitioning to node $n'$ sampled from $p_{n'|no'}$ and so on. The value of a controller can be computed by solving a linear system: $V_{ns} = \sum_a p_{a|n} [r_{as} + \gamma \sum_{s'o'n'} p_{s'|as} p_{o'|s'a} p_{n'|no'} V_{n's'}] \; \forall ns$. The value at a given belief $b$ is then $V^\pi(b) = \sum_n \sum_s b_s p_n V_{ns}$.

### 2.3 Policy Optimization

Several techniques have been proposed to optimize controllers of a given size, including gradient ascent [9], stochastic local search [2], bounded policy iteration [14], non-convex quadratically constrained optimization [1] and likelihood maximization [18]. We briefly describe the last technique since we will use it in Sect. 4.

Toussaint et al. [18] recently proposed to convert POMDPs into equivalent dynamic Bayesian networks (DBNs) by normalizing the rewards and to optimize a policy by maximizing the likelihood of the normalized rewards. Let $\tilde{R}$ be a binary variable corresponding to normalized rewards. The reward function $r_{as}$ is then replaced by a reward distribution $p_{\tilde{r}|sat} = \Pr(\tilde{R} = \tilde{r} \mid A_t = a, S_t = s, T = t)$ that assigns probability $r_{as}/(r_{max} - r_{min})$ to $\tilde{R} = 1$ and $1 - r_{as}/(r_{max} - r_{min})$ to $\tilde{R} = 0$ ($r_{min} = \min_{as} r_{as}$ and $r_{max} = \max_{as} r_{as}$). An additional time variable $T$ is introduced to simulate the discount factor and the summation of rewards. Since a reward is normally discounted by a factor $\gamma^t$ when earned $t$ time steps in the future, the prior $p_t = \Pr(T = t)$ is set to $\gamma^t(1-\gamma)$ where the factor $(1-\gamma)$ ensures that $\sum_{t=0}^\infty p_t = 1$. The resulting dynamic Bayesian network is illustrated in Fig. 1. It can be thought of as a mixture of finite processes of length $t$ with a 0-1 reward $\tilde{R}$ earned at the end of the process. The nodes $N_t$ encode the internal memory of the controller. Given the controller distributions $p_n$, $p_{a|n}$ and $p_{n'|no'}$, it is possible to evaluate the controller by computing the likelihood of $\tilde{R} = 1$. More precisely, $V^\pi(p_s) = (\Pr(\tilde{R}=1) - r_{min})/[(r_{max} - r_{min})(1-\gamma)]$.

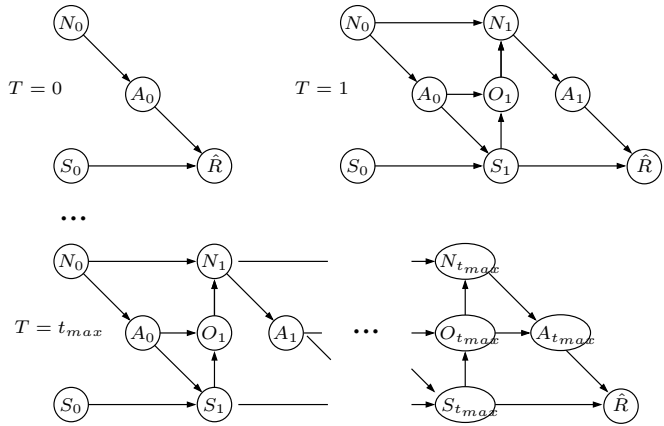

Figure 1: POMDP represented as a mixture of finite DBNs. For an infinite horizon, a large enough $t_{max}$ can be selected at runtime to ensure that the approximation error is small.

Optimizing the policy can be framed as maximizing the likelihood of $\tilde{R} = 1$ by varying the distributions $p_n$, $p_{a|n}$ and $p_{n'|no'}$ encoding the policy. Toussaint et al. use the expectation-maximization (EM) algorithm. Since EM is guaranteed to increase the likelihood at each iteration, the controller's value increases monotonically. However, EM is not guaranteed to converge to a global optimum. An important advantage of the EM algorithm is its simplicity both conceptually and computationally. In particular, the computation consists of inference queries that can be computed using a variety of exact and approximate algorithms.

## 3 Hierarchical Modeling

While optimizing a bounded controller allows an effective search in the space of bounded policies, such an approach is clearly suboptimal since the optimal controller of many problems grows doubly exponentially with the planning horizon and may be infinite for infinite horizons. Alternatively, hierarchical representations permit the representation of structured policies with exponentially fewer parameters. Several approaches were recently explored to model and learn hierarchical structures in POMDPs. Pineau et al. [13] sped up planning by exploiting a user specified action hierarchy. Hansen et al. [7] proposed hierarchical controllers and an alternative planning technique that also exploits a user specified hierarchy. Charlin et al. [4] proposed recursive controllers (which subsume hierarchical controllers) and an approach that discovers the hierarchy while optimizing a controller. We briefly review recursive controllers in Sect. 3.1 since we will empirically compare our approach to the non-convex optimization techniques used to optimize recursive controllers. In another line of research, Murphy and Paskin [10] proposed to model hierarchical hidden Markov models (HMMs) with a dynamic Bayesian network (DBN). Theocharous et al. [17] also used DBNs to model hierarchical POMDPs. We briefly review this DBN encoding in Sect. 3.2 since we will use it in our approach to model factored controllers.

### 3.1 Recursive Controllers

A recursive controller [4] consists of a recursive automaton with concrete nodes $n$ and abstract nodes $\bar{n}$. Abstract nodes call a subcontroller before selecting an action. A controller is said to be recursive when it can call itself, essentially encoding an infinite hierarchy. Formally, a recursive controller is parametrized by an action selection distribution for each node (e.g., $p_{a|n}$ and $p_{a|\bar{n}}$), a successor node distributions for each node (e.g., $p_{n'|no'}$ and $p_{n'|\bar{n}o'}$) and a child node distribution for each abstract node (e.g., $p_{n'|\bar{n}}$)[1]. Execution of a recursive controller is performed by executing the action selected by each node visited and continuing to the successor node selected by the observation made. However, when an abstract node is visited, before executing the action selected, its subcontroller is called and started in the child node selected by the child node distribution. A subcontroller returns control to its parent node when a special end node is reached. Charlin et al. [4] show that optimizing a recursive controller with a fixed number of concrete and abstract nodes can be framed as a non-convex quartically constrained optimization problem. The hierarchical structure is discovered as the controller is optimized since the variables of the optimization problem include the child node distributions which implicitly encode the hierarchy. Three techniques based on a general non-linear solver, a mixed-integer non-linear approximation and a form of bounded hierarchical policy iteration are experimented with, but do not scale beyond toy problems. Furthermore, Charlin et al. do not demonstrate whether searching in the space of hierarchical controllers can speed up planning. Although it is clear that planning is simplified when a hierarchy is given a priori since the policy space is reduced, it is not clear that hierarchy discovery is beneficial since the policy space remains the same while the parameter space changes. In Sect. 5, we demonstrate that hierarchy discovery can be beneficial when a simple hierarchical policy of high value exists.

---
[1]The $p_{a|n}$ and $p_{n'|no'}$ distributions are combined in one distribution $p_{n'a|no'}$ in [14]

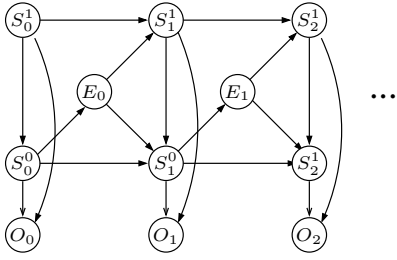

Figure 2: DBN encoding of a 2-level hierarchical HMM.

### 3.2 Hierarchical HMMs

Murphy and Paskin [10] proposed to model hierarchical hidden Markov models (HMMs) as dynamic Bayesian networks (DBNs). The idea is to convert a hierarchical HMM of $L$ levels into a dynamic Bayesian network of $L$ state variables, where each variable encodes abstract states at the corresponding level. Here, abstract states can only call sub-HMMs at the previous level. Fig. 2 illustrates a two-level hierarchical HMMs encoded as a DBN. The state variables $S_t^l$ are indexed by the time step $t$ and the level $l$. The $E_t$ variables indicate when a base-level sub-HMM has ended, returning its control to the top level HMM. The top-level abstract state transitions according to the top HMM, but only when the exit variable $E_t$ indicates that the base-level concrete state is an exit state. The base-level concrete state transitions according to the base-level HMM. When an exit state is reached, the next base-level state is determined by the next top-level abstract state. Factored HMMs subsume hierarchical HMMs in the sense that there exists an equivalent factored HMM for every hierarchical HMM. In Sect. 4.1, we will use a similar technique to convert hierarchical controllers into factored controllers.

## 4 Factored Controllers

We propose to combine the DBN encoding techniques of Murphy et al. [10] and Toussaint et al. [18] to convert a POMDP with a hierarchical controller into a mixture of DBNs. The hierarchy and the controller are simultaneously optimized by maximizing the reward likelihood of the DBN. We also consider factored controllers which subsume hierarchical controllers.

### 4.1 DBN Encoding

Fig. 3a illustrates two consecutive slices of one DBN in the mixture (rewards are omitted) for a three-level hierarchical controller. Consider a POMDP defined by the tuple $\langle \mathcal{S}, \mathcal{A}, \mathcal{O}, p_s, p_{s'|as}, p_{o'|s'a}, r_{as} \rangle$ and a three-level hierarchical (non-recursive) controller defined by the tuple $\langle p_{a|n^l}, p_{n^{l-1}|n^l}, p_{n'^l|n^l o'} \rangle \; \forall l$. The conditional probability distributions of the mixture of DBNs (denoted by $\hat{p}$) are:

- transition distribution: $\hat{p}_{s'|as} = p_{s'|as}$

- observation distribution: $\hat{p}_{o'|s'a} = p_{o'|s'a}$

- reward distribution:
  $\hat{p}_{\tilde{r}|as} = (r_{as} - r_{min})/(r_{max} - r_{min})$

- mixture distribution: $\hat{p}_t = (1-\gamma)\gamma^t$

- action distribution: $\hat{p}_{a|n^0} = p_{a|n^0}$

- base level node distribution: $\hat{p}_{n'^0|n^0 n'^1 o' e^0}$
  $= \begin{cases} p_{n'^0|n'^1} & \text{if } e^0 = exit \\ p_{n'^0|o'n^0} & \text{otherwise} \end{cases}$

- middle level node distribution: $\hat{p}_{n'^1|n^1 n'^2 o' e^0 e^1}$
  $= \begin{cases} p_{n'^1|n'^2} & \text{if } e^1 = exit \\ p_{n'^1|o'n^1} & \text{if } e^0 = exit \text{ and } e^1 \neq exit \\ \delta_{n'^1 n^1} & \text{otherwise} \end{cases}$

- top level node distribution: $\hat{p}_{n'^2|o'n^2 e^1}$
  $= \begin{cases} p_{n'^2|o'n^2} & \text{if } e^1 = exit \\ \delta_{n'^2 n^2} & \text{otherwise} \end{cases}$

- base-level exit distribution: $\hat{p}_{e^0|n^0}$
  $= \begin{cases} 1 & \text{if } n^0 \text{ is an end node} \\ 0 & \text{otherwise} \end{cases}$

- middle-level exit distribution: $\hat{p}_{e^1|n^1 e^0}$
  $= \begin{cases} 1 & \text{if } e^0 = exit \text{ and } n^1 \text{ is an end node} \\ 0 & \text{otherwise} \end{cases}$

While the $E_t^l$ variables help clarify when the end of a sub-controller is reached, they are not necessary. Eliminating them yields a simpler DBN illustrated in Fig. 3b. The conditional probability distributions of the $N_t^l$ variables become:

- base level node distribution: $\hat{p}_{n'^0|n^0 n'^1 o'}$
  $= \begin{cases} p_{n'^0|n'^1} & \text{if } n^0 \text{ is an end node} \\ p_{n'^0|o'n^0} & \text{otherwise} \end{cases}$

- middle level node distribution: $\hat{p}_{n'^1|n^1 n'^2 o'}$
  $= \begin{cases} p_{n'^1|n'^2} & \text{if } n^1 \text{ and } n^0 \text{ are end nodes} \\ p_{n'^1|o'n^1} & \text{if } n^0 \text{ is an end node, but not } n^1 \\ \delta_{n'^1 n^1} & \text{otherwise} \end{cases}$

- top level node distribution: $\hat{p}_{n'^2|n^2 o' e^1}$
  $= \begin{cases} p_{n'^2|n^2 o'} & \text{if } n^1 \text{ and } n^0 \text{ are end nodes} \\ \delta_{n'^2 n^2} & \text{otherwise} \end{cases}$

Note that ignoring the above constraints in the conditional distributions yields a *factored controller* that is more flexible than a hierarchical controller since the conditional probability distributions of the $N_t^l$ variables do not have to follow the structure imposed by a hierarchy

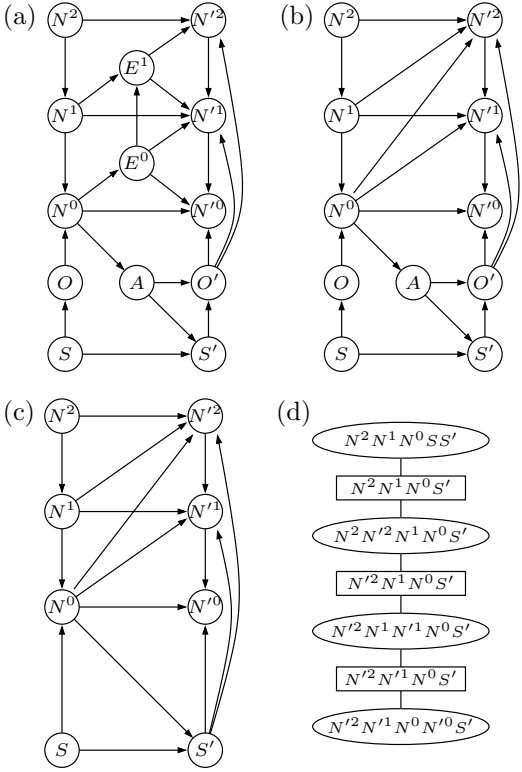

Figure 3: (a) Two slices of the DBN encoding the hierarchical POMDP controller. (b) A version where exit variables are eliminated. (c) Variables $O$ and $A$ are eliminated. (d) The corresponding junction tree (or rather chain) for inference.

## 4.2 Maximum Likelihood Estimation

Following Toussaint et al.'s technique [18], we optimize a factored controller by maximizing the reward likelihood. Since the policy parameters are conditional probability distributions of the DBN, the EM algorithm can be used to optimize them. Computation alternates between the E and M steps below. We denote by $n^{top}$ and $n^{base}$ the top and base nodes in a given time slice. We also denote by $\phi(V)$ and $\phi(v)$ the parents of $V$ and a configuration of the parents of $V$.

**E-step:** expected frequency of the hidden variables
$E_{n^{top}} = \Pr(N_0^{top} = n^{top} | \tilde{R}=1)$
$E_{an^{base}} = \sum_t \Pr(A_t=a, N_t^{base}=n^{base} | \tilde{R}=1)$
$E_{n'^l \phi(n'^l)} =$
$\quad \sum_t \Pr(N_{t+1}^l = n'^l, \phi(N_{t+1}^l) = \phi(n_{t+1}^l) | \tilde{R}=1) \; \forall l$

**M-step:** relative frequency computation
$p_{n^{top}} = E_{n^{top}} / \sum_{n^{top}} E_{n^{top}}$
$p_{a|n^{base}} = E_{an^{base}} / \sum_a E_{an^{base}}$
$p_{n'^l | \phi(n'^l)} = E_{n'^l \phi(n'^l)} / \sum_{n'^l} E_{n'^l \phi(n'^l)} \; \forall l$

### 4.2.1 Parameter initialization

W.l.o.g. we initialize the start node $N_0^{top}$ of the top layer to be the first node (i.e., $\Pr(N_0^{top}=1)=1$). The node conditional distributions $p_{n'^l | \phi(n'^l)}$ are initialized randomly as a mixture of three distributions:

$$p_{n'^l | \phi(n'^l)} \propto c_1 + c_2 \mathcal{U}_{n'^l \phi(n'^l)} + c_3 \delta_{n'^l n^l}$$

The mixture components are a uniform distribution, a random distribution $\mathcal{U}_{\phi(n'^l)}$ (an array of uniform random numbers in $[0,1]$), and a term enforcing $n^l$ to stay unchanged. For the node distributions at the base level we choose $c_1=1, c_2=1, c_3=0$ and for all other levels we choose $c_1=1, c_2=1, c_3=10$. Similarly we initialize the action probabilities as

$$p_{a|n^{base}} \propto c_1 + c_2 \mathcal{U}_{an^{base}} + c_3 \delta_{a(n^{base} \% a)}$$

with $c_1=1, c_2=1, c_3=100$, where the last term enforces each node $n^{base}=i$ to be associated with action $a=i\%a$.

### 4.2.2 E-step

To speed up the computation of the inference queries in the E-step, we compute intermediate terms using a forward-backward procedure. Let $t_{max}$ be the largest value of $T$, then a simple scheme that answers each query separately takes $O(t_{max}^2)$ time since there are $O(t_{max})$ queries and each query takes $O(t_{max})$ time to run over the entire network. However, since part of the computation is duplicated in several queries, it is possible to compute intermediate terms $\alpha$ and $\beta$ in $O(t_{max})$ time from which each expectation can be computed in constant time (w.r.t. $t_{max}$). To simplify notation, $\mathbf{N}$ and $\mathbf{n}$ denote all the nodes and their joint configuration in a given time slice.

**Forward term:** $\alpha_{\mathbf{n}s}^t = \Pr(\mathbf{N}_t = \mathbf{n}, S_t = s)$
$\alpha_{\mathbf{n}s}^0 = p_{\mathbf{n}} p_s$
$\alpha_{\mathbf{n}'s'}^t = \sum_{\mathbf{n},s} \alpha_{\mathbf{n}s}^{t-1} p_{\mathbf{n}'s'|\mathbf{n}s}$

**Backward term:** $\beta_{\mathbf{n}s}^\tau = \Pr(\tilde{R}=1 | \mathbf{N}_{t-\tau} = \mathbf{n}, S_{t-\tau} = s, T=t)$
$\beta_{\mathbf{n}s}^0 = \sum_a p_{a|\mathbf{n}} r_{as}$
$\beta_{\mathbf{n}s}^\tau = \sum_{\mathbf{n}',s'} p_{\mathbf{n}'s'|\mathbf{n}s} \beta_{\mathbf{n}'s'}^{\tau-1}$

To fully take advantage of the structure of the DBN, we first marginalize the DBN w.r.t. the observations and actions to get the DBN in Fig. 3c. This 2-slice DBN corresponds to the joint transition distribution $p_{\mathbf{n}'s'|\mathbf{n}s}$ used in the above equations. Then we compile this 2-slice DBN into the junction tree (actually junction chain) given in Fig. 3d.

Let $\beta_{\mathbf{n}s} = \sum_\tau \Pr(T=\tau) \beta_{\mathbf{n}s}^\tau$ and $\alpha_{\mathbf{n}s} = \sum_t \Pr(T=t) \alpha_{\mathbf{n}s}^t$, then the last two expectations of the E-step

can be computed as follows:[2]

$$E_{an^{base}} \propto \sum_{s,\mathbf{n}-\{n^{base}\}} \alpha_{\mathbf{ns}} p_{a|n^{base}} [r_{as} + \gamma \sum_{s',o',\mathbf{n'}} p_{s'|as} p_{o'|s'a} p_{\mathbf{n'}|o'\mathbf{n}} \beta_{\mathbf{n's'}}]$$
$$E_{n'^l \phi(n'^l)} \propto \sum_{s,s',a,\mathbf{n}-\phi(n'^l),n'^{-l}} \alpha_{\mathbf{ns}} p_{a|n^{base}} p_{s'|as} p_{o'|s'a} p_{\mathbf{n'}|o'\mathbf{n}} [r_{as} + \gamma \beta_{\mathbf{n's'}}] \; \forall l$$

### 4.2.3 M-step

The standard M-step adjusts each parameter $p_{v|\phi(v)}$ by normalizing the expectations computed in the E-step, i.e., $p_{v|\phi(v)}^{\text{new}} \propto E_{v\phi(v)}$. To speed up convergence, we instead use a variant that performs a soften *greedy M-step*. In the greedy M-step, each parameter $p_{v|\phi(v)}^{\text{new}}$ is greedily set to 1 when $v = \text{argmax}_{\bar{v}} f_{\bar{v}\phi(\bar{v})}$ and 0 otherwise, where $f_{v\phi(v)} = E_{v\phi(v)}/p_{v|\phi(v)}^{\text{old}}$. The greedy M-step can be thought of as the limit of an infinite sequence of alternating partial E-step and standard M-step where the partial E-step keeps $f$ fixed. The combination of a standard M-step with this specific partial E-step updates $p_{v|\phi(v)}$ by a multiplicative factor proportional to $f_{v\phi(v)}$. In the limit, the largest $f_{v\phi(v)}$ ends up giving all the probability to the corresponding $p_{v|\phi(v)}$. EM variants with certain types of partial E-steps ensure monotonic improvement of the likelihood when the hidden variables are independent [11]. This is not the case here, however by softening the greedy M-step we can still obtain monotonic improvement most of the time while speeding up convergence. We update $p_{v|\phi(v)}$ as follows:

$$v^* = \underset{v}{\text{argmax}} \; f_{v\phi(v)}$$
$$p_{v|\phi(v)}^{\text{new}} \propto p_{v|\phi(v)}^{\text{old}}[\delta_{vv^*} + c + \epsilon] \; .$$

For $c = 0$ and $\epsilon = 0$ this is the greedy M-step. We use $c = 3$ which softens (shortens) the step and improves convergence. Furthermore, adding small Gaussian noise $\epsilon \sim \mathcal{N}(0, 10^{-3})$ helps to escape local minima.

### 4.2.4 Complexity

For a flat controller, the number of parameters (neglecting normalization) is $|\mathcal{O}||\mathcal{N}|^2$ for $p_{n'|o'n}$ and $|\mathcal{A}||\mathcal{N}|$ for $p_{a|n}$. The complexity of the forward (backward) procedure is $O(t_{max}(|\mathcal{N}||\mathcal{S}|^2 + |\mathcal{N}|^2|\mathcal{S}|))$ where the two terms correspond to the size of the two cliques for inference in the 2-slice DBN after $O$ and $A$ are eliminated. The complexity of computing the expectations from $\alpha$ and $\beta$ is $O(|\mathcal{N}||\mathcal{A}|(|\mathcal{S}|^2 + |\mathcal{S}||\mathcal{O}|) + |\mathcal{N}|^2|\mathcal{S}||\mathcal{O}|)$, which corresponds to the clique sizes of the 2-slice DBN including $O$ and $A$.

In comparison, 2-level hierarchical and factored controllers with $|\mathcal{N}^{top}| = |\mathcal{N}^{base}| = |\mathcal{N}|^{0.5}$ nodes at each

---

[2]The first expectation of the E-step does not need to be computed since $\Pr(N_0^{top} = 1) = 1$.

Table 1: Number of parameters and computational complexity for the flat controller with $|\mathcal{N}|$ nodes and a 2-layer factored controller with $|\mathcal{N}^{top}| = |\mathcal{N}^{base}| = |\mathcal{N}|^{0.5}$ nodes.

| # parameters | |
|---|---|
| flat | $|\mathcal{O}||\mathcal{N}|^2 + |\mathcal{A}||\mathcal{N}|$ |
| fact. | $2|\mathcal{O}||\mathcal{N}|^{1.5} + |\mathcal{A}||\mathcal{N}|^{0.5}$ |
| forward-backward complexity | |
| flat | $O(t_{max}(|\mathcal{N}||\mathcal{S}|^2 + |\mathcal{N}|^2|\mathcal{S}|))$ |
| fact. | $O(t_{max}(|\mathcal{N}||\mathcal{S}|^2 + |\mathcal{N}|^{1.5}|\mathcal{S}|))$ |
| expectation complexity | |
| flat | $O(|\mathcal{N}||\mathcal{A}|(|\mathcal{S}|^2 + |\mathcal{S}||\mathcal{O}|) + |\mathcal{N}|^2|\mathcal{S}||\mathcal{O}|)$ |
| fact. | $O(|\mathcal{N}||\mathcal{A}|(|\mathcal{S}|^2 + |\mathcal{S}||\mathcal{O}|) + |\mathcal{N}|^{1.5}|\mathcal{O}||\mathcal{S}| + |\mathcal{N}|^2|\mathcal{O}|)$ |

level have fewer parameters and a smaller complexity, but also a smaller policy space due to the structure imposed by the hierarchy/factorization. While there is a tradeoff between policy space and complexity, hierarchical and factored controllers are often advantageous in practice since they can find more quickly a good hierarchical/factored policy when there exists one.

A 2-level factored controller with $|\mathcal{N}|^{0.5}$ nodes at each level has $2|\mathcal{O}||\mathcal{N}|^{1.5}$ parameters for $p_{n'^{top}|o'n^{base}n^{top}}$ and $p_{n'^{base}|n'^{top}o'n^{base}}$, and $|\mathcal{A}||\mathcal{N}|^{0.5}$ parameters for $p_{a|n^{base}}$. The complexity of the forward (backward) procedure is $O(t_{max}(|\mathcal{N}||\mathcal{S}|^2 + |\mathcal{N}|^{1.5}|\mathcal{S}|))$ and the complexity of computing the expectations is $O(|\mathcal{N}||\mathcal{A}|(|\mathcal{S}|^2 + |\mathcal{S}||\mathcal{O}|) + |\mathcal{N}|^{1.5}|\mathcal{O}||\mathcal{S}| + |\mathcal{N}|^2|\mathcal{O}|)$. A 2-level hierarchical controller is further restricted and therefore has fewer parameters, but the same time complexity.

## 5 Experiments

We first compared the performance of the maximum likelihood (ML) approach to previous optimization-based approaches from [4]. Table 2 summarizes the results for 2-layer controllers with certain combinations of $|\mathcal{N}^{base}|$ and $|\mathcal{N}^{top}|$. The problems include paint, shuttle and 4x4 maze (previously used in [4]) and three additional problems: chain-of-chains (described below), hand-washing (reduced version from [8]) and cheese-taxi (variant from [12]). On the first three problems, ML reaches the same values as the previous optimization-based approaches, but with larger controllers. We attribute this to EM's weaker ability to avoid local optima than the optimization-based approaches. However, the optimization-based approaches run out of memory on the last three problems (memory needs exceed 2 Gb of RAM), while ML scales gracefully (as analyzed in Sect. 4.2.4). ML approach demonstrates that hierarchy discovery can be tackled with tractable algorithms. We also report the values reached with a state of the art point-based value

Table 2: $V^*$ denotes optimal values (with truncated trajectories) [3] except for handwashing and cheese-taxi where we show the optimal value of the equivalent fully-observable problem. HSVI2 found a solution in less than 1s for every problem except handwashing where the algorithm was halted after 12 hours of computation. The ML approach optimizes a factored controller for 200 EM iterations with a planning horizon of $t_{max}=100$. (5,3) nodes means $|\mathcal{N}^{base}|=5$ and $|\mathcal{N}^{top}|=3$. For cheese-taxi, we get a maximum value of 2.25. N/A indicates that the solver did not complete successfully. All tests are done on a dual-core x64 processor @2.2GHz.

| Problem | $|\mathcal{S}|,|\mathcal{A}|,|\mathcal{O}|$ | $V^*$ | HSVI2 $V$ | Best results from [4] | | | ML approach (avg. over 10 runs) | | |
|---|---|---|---|---|---|---|---|---|---|
| | | | | nodes | t(s) | $V$ | nodes | t(s) | $V$ |
| paint | 4, 4, 2 | 3.28 | 3.29±0.04 | (1,3) | <1 | 3.29 | (5,3) | 0.96±0.3 | 3.26±0.004 |
| shuttle | 8, 3, 5 | 32.7 | 32.9±0.8 | (1,3) | 2 | 31.87 | (5,3) | 2.81±0.2 | 31.6±0.5 |
| 4x4 maze | 16, 4, 2 | 3.7 | 3.75±0.1 | (1,2) | 30 | 3.73 | (3,3) | 2.8±0.8 | 3.72±8e−5 |
| chain-of-chains | 10, 4, 1 | 157.1 | 157.1±0 | (3,3) | 10 | 0.0 | (10,3) | 6.4±0.2 | 151.6±2.6 |
| handwashing | 84, 7, 12 | ⩽1052 | N/A | | | N/A | (10,5) | 655±2 | 984±1 |
| cheese-taxi | 33, 7, 10 | ⩽5.3 | 2.53±0.3 | | | N/A | (10,3) | 311±14 | −9±11(2.25*) |

iteration method (HSVI2 [15]).

The next question is whether there are computational savings when automatically discovering a hierarchy. Recall that previous work has shown that policy optimization is simplified when a hierarchy is known a priori since the space of policies is restricted. The next experiment demonstrates that policy optimization while discovering a hierarchy can be done faster and/or yield higher value when there exists good hierarchical policies. Table 3 compares the performance when optimizing flat, hierarchical and factored controllers on chain-of-chains, hand-washing and cheese-taxi. Here, the factored and hierarchical controllers have two levels and correspond respectively to the DBNs in Fig. 3(a) and 3(b).[3] The x-axis is the number of nodes for flat controllers and the product of the number of nodes at each level for hierarchical and factored controllers. Taking the product is justified by the fact that the equivalent flat controllers of some hierarchical/factored controllers require that many nodes. The graphs in the right column of Table 3 demonstrate that hierarchical and factored controllers can be optimized faster, confirming the analysis done in Sect. 4.2.4. There is no difference in computational complexity between the strictly hierarchical and unconstrained factored architectures. Recall however that the efficiency gains of the hierarchical and factored controllers are obtained at the cost of a restricted policy space. Nevertheless, the graphs in the left column of Table 3 suggest that hierarchical/factored controllers can still find equally good policies when there exist one. Factored controllers are generally the most robust. With a sufficient number of nodes, they find the best policies on all three problems. Note that factored and hierarchical controllers need at least a number of nodes equal to the number of actions in the base layer in order to represent a policy that uses all actions.

[3]Factored controllers are hierarchical controllers where the restrictions imposed by the $E_t$ variables are removed.

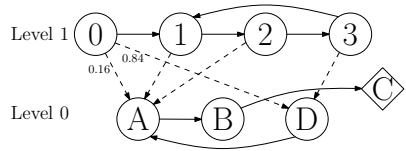

Figure 4: Hierarchical controller learnt for the chain-of-chains. The diamond indicates an exit node, for which $\hat{p}_{e^0|n^0} = 1$.

This explains why hierarchical and factored controllers with less than 4 base nodes (for chain-of-chains) and 7 base nodes (for hand-washing and cheese-taxi) do poorly. The optimization of flat controllers tend to get stuck in local optima if too many nodes are used. Comparing the unconstrained factored architecture versus hierarchical, we find that the additional constraints in the hierarchical controller make the optimization problem harder although there are less parameters to optimize. As a result, EM gets stuck more often in local optima.

We also examine whether learnt hierarchies make intuitive sense. Good policies for the cheese-taxi and hand-washing problems can often be represented hierarchically, however the hierarchical policies found didn't match hierarchies expected by the authors. Since these are non-trivial problems for which there may be many ways to represent good policies in a hierarchical fashion that is not intuitive, we designed the chain-of-chains problem, which is much simpler to analyze. The optimal policy of this problem consists of executing $n$ times the same chain of $n$ actions followed by a submit action to earn the only reward. The optimal policy requires $n^2+1$ nodes for flat controllers and $n+1$ nodes at each level for hierarchical controllers. For $n=3$, ML found a hierarchical controller of 4 nodes at each level, illustrated in Fig. 4. The controller starts in node 0. Nodes at level 1 are abstract and descend into concrete nodes at level 0 by following the dashed edges. Control is returned to level 1 when an end node (denoted by a diamond) is reached. Here, the optimal policy is to do A-B-C three times followed by D. Hence a natural hierarchy would abstract A-B-C and D into separate subcontrollers. While the controller in Fig. 4 is not completely optimal (the vertical transition from abstract node 0 should have probability 1 of reaching node A), it found an equivalent, but less intuitive abstraction by having subcontrollers that do A-B-C and D-A-B-C. This suggests that for real-world problems there will be many valid abstractions that are not easily interpretable by humans and the odds that an automated procedure finds an intuitive hierarchy without any additional guidance are slim.

Table 3: *Left:* The reached values depending on the number of nodes in the controller. For the factored and hierarchical controller we indicate the number of nodes in both layers (e.g., (5,3) means $|\mathcal{N}^{base}| = 5$ and $|\mathcal{N}^{top}| = 3$) and plot the data point at $|\mathcal{N}^{base}||\mathcal{N}^{top}|$ on the x-axis. For instance, in the case of handwashing we see how the performance depends critically on $|\mathcal{N}^{base}|$. *Right:* The optimization time. In all cases, 200 EM iterations are performed with a planning horizon of $t_{max} = 100$. The results for each controller are the average of 10 runs with error bars of $\pm 1$ standard deviation.

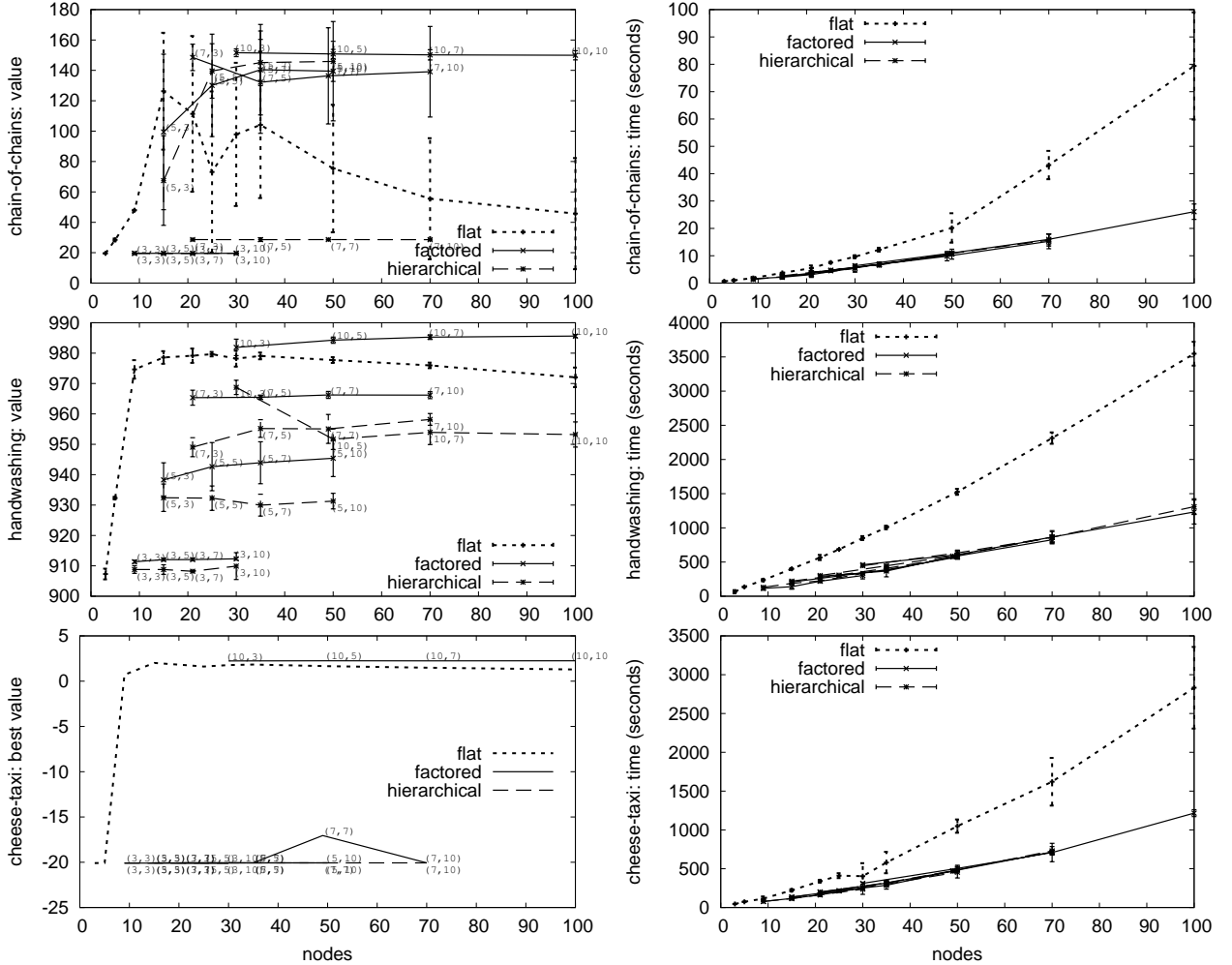

## 6   Conclusion

The key advantage of maximum likelihood is that it can exploit the factored structure in a controller architecture. This facilitates hierarchy discovery when the hierarchical structure of the controller is encoded into a corresponding dynamic Bayesian network (DBN). Our complexity analysis and the empirical run time analysis confirm the favorable scaling. In particular, we solved problems like handwashing and cheese-taxi that could not be solved with the previous approaches in [4]. Compared to flat controllers, factored controllers are faster to optimize and less sensitive to local optima when they have many nodes. Our current implementation does not exploit any factored structure in the state, action and observation space, however we envision that a factored implementation would naturally scale to large factored POMDPs.

For the chain-of-chains problem, maximum likelihood finds a valid hierarchy. For other problems like handwashing, there might be many hierarchies and the one found by our algorithm is usually hard to interpret. We cannot expect our method to find a hierarchy that is human readable. Interestingly, although the strictly hierarchical architectures have less parameters to optimize, they seem to be more susceptible to local optima as compared to a factored but otherwise unconstrained controller. Future work will investigate various heuristics to escape local optima during optimization.

In this paper we made explicit assumptions about the structure – we prefixed the structure of the DBN to mimic a strict hierarchy or a level-wise factorization and we fixed the number of nodes in each level. However, the DBN framework allows us to build on existing methods for structure learning of graphical models. A promising extension would be to use such structure learning techniques to optimize the factored structure of the controller. Since the computational complexity for evaluating (training) a single structure is reasonable, techniques like MCMC could sample and evaluate a variety of structures. This variety might also help to circumvent local optima, which currently define the most dominant limit of our approach.

**Acknowledgments**

Part of this work was completed while Charlin was at the University of Waterloo. Toussaint acknowledges support by the German Research Foundation (DFG), Emmy Noether fellowship TO 409/1-3. Poupart and Charlin were supported by grants from the Natural Sciences and Engineering Research Council of Canada, the Canada Foundation for Innovation and the Ontario Innovation Trust.